\documentclass[10pt,twocolumn,letterpaper]{article}

\usepackage{iccv}
\usepackage{times}
\usepackage{epsfig}
\usepackage{graphicx}
\usepackage{amsmath}
\usepackage{amssymb}
\usepackage{verbatim}
\usepackage{subcaption}

\usepackage{mathtools}
\usepackage{multirow}
\usepackage{authblk}
\usepackage[symbol]{footmisc}
\usepackage{multirow}
\usepackage{ctable} 
\usepackage{floatrow}
\usepackage{algorithm}
\usepackage{algorithmic}
\usepackage{multirow}
\usepackage{tabularx}
\usepackage{blindtext}

\usepackage[export]{adjustbox}

\usepackage[pagebackref=true,breaklinks=true,letterpaper=true,colorlinks,bookmarks=false]{hyperref}

\iccvfinalcopy 

\ificcvfinal\fi

\definecolor{grigoriscolor}{RGB}{219, 48, 122}

\DeclareMathAlphabet\mathbfcal{OMS}{cmsy}{b}{n}

\begin{document}

\title{Poly-NL: Linear Complexity Non-local Layers with Polynomials}
\author{
Francesca Babiloni\textsuperscript{1}, Ioannis Marras\textsuperscript{1}, Filippos Kokkinos\textsuperscript{3}, Jiankang Deng\textsuperscript{2}, Grigorios Chrysos\textsuperscript{4}, Stefanos Zafeiriou\textsuperscript{2} \\
\textsuperscript{1}Huawei Noah's Ark Lab \quad \textsuperscript{2}Imperial College London \quad \textsuperscript{3}University College London  \quad \\
\textsuperscript{4}École Polytechnique Fédérale de Lausanne  \quad 
}
\maketitle
\begin{abstract}
Spatial self-attention layers, in the form of Non-Local blocks, introduce long-range dependencies in Convolutional Neural Networks by computing pairwise similarities among all possible positions. Such pairwise functions underpin the effectiveness of non-local layers, but also determine a complexity that scales quadratically with respect to the input size both in space and time. This is a severely limiting factor that practically hinders the applicability of non-local blocks to even moderately sized inputs. Previous works focused on reducing the complexity by modifying the underlying matrix operations, however in this work we aim to retain full expressiveness of non-local layers while keeping complexity linear. We overcome the efficiency limitation of non-local blocks by framing them as special cases of 3rd order polynomial functions. 
This fact enables us to formulate novel fast Non-Local blocks, capable of reducing the complexity from quadratic to linear with no loss in performance, by replacing any direct computation of pairwise similarities with element-wise multiplications. The proposed method, which we dub as ``Poly-NL", is competitive with state-of-the-art performance across image recognition, instance segmentation, and face detection tasks, while having considerably less computational overhead.
\end{abstract}

\section{Introduction\vspace{-0.1cm}}
Convolutional Neural Networks (CNNs) have led to a revolution in machine learning, and, specifically, are currently the undisputed state of the art in computer vision on various tasks. Nonetheless, CNNs, even if composed by a deep stack of convolutional operators, have a limited receptive field~\cite{luo2017understanding}, which makes the crucial long-range dependencies hard to capture.

Recent work on spatial self-attention ameliorated this issue with a novel set of modules for neural networks~\cite{wang2018non, vaswani2017attention}. These blocks extract non-local interactions among all spatial positions of the input and weight them with a set of learnable parameters. Passing through a Non-local block, each input position takes into account the contribution of all the others, scaled by their similarity with a given reference. These blocks introduce the possibility to reason about the whole space in one glance and make non-local behavior easier to be captured by the network. Inserting Non-local blocks in neural architectures has been proven very effective~\cite{Bello_2019_ICCV,dai2019second,wang2019edvr,ott2018scaling,ramachandran2019stand, mohamed2019transformers}, but the computation of a similarity score for each pair of points scales quadratically with the number of spatial positions. As such, the expensive computational and storage complexity makes non-local blocks impractical to compute even upon moderately sized input. 

Recent works tackle such limitation via an efficient computation of the similarity matrix~\cite{zhang2019latentgnn, liu2019spatially, shen2021efficient} but miss to provide a theoretical overview of the Non-local block formulation. 
In this work, we build upon the aforementioned line of research, and revisit Non-local layers under the lens of polynomials, framing them as special cases of $3^{rd}$ order polynomials. Powered by this intuition, we derive an efficient version of Non-local neural networks, \textit{Poly-NL} which takes into account long-range dependencies without the need to compute explicitly any pairwise similarity. Poly-NL layers perform computations using the same set of interactions as the Non-local block of~\cite{wang2018non}, and at the same time reduce the overall complexity drastically from $O(N^2)$ to $O(N)$ with no loss in performance.

In this work, we link polynomials and the Non-Local layer. Our goal is to efficiently extract high-order interactions from the input while capturing long-range spatial dependencies. Thus, our contribution can be summarized as follows:
\begin{itemize}
    \item We bridge the formulations between high-order polynomials and non-local attention. In particular, we prove that self-attention (in the form of Non-local blocks) can be seen as a particular case of general $3^{rd}$ order polynomials.

    \item We propose "Poly-NL" a novel building block for neural networks, which can be seen as polynomials of the input matrix. In particular, we propose an alternative Non-local block that reduce complexity from quadratic to linear with respect to the spatial dimensions. 
    
    \item We showcase the efficiency and the effectiveness of our method across a range of tasks: image recognition, instance segmentation, and face detection.
    
\end{itemize}

\begin{figure*}[t]
  \subfloat[\centering]{\label{fig:big_c}\includegraphics[width=.40\linewidth, valign=c ]{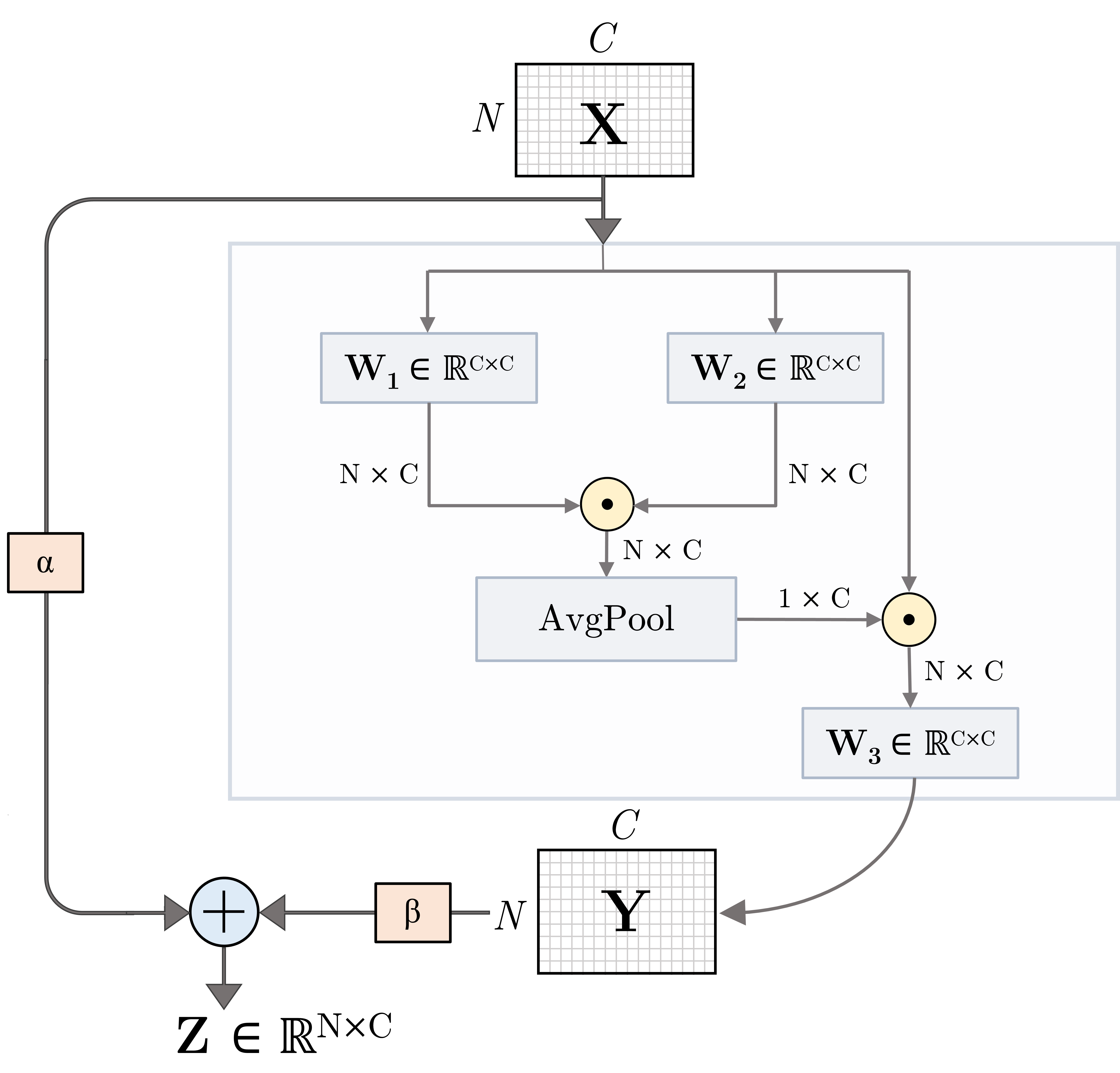}}
  \hspace*{0.5cm}
  \subfloat[\centering]{\label{fig:big_s}\includegraphics[width=.55\linewidth, valign=c]{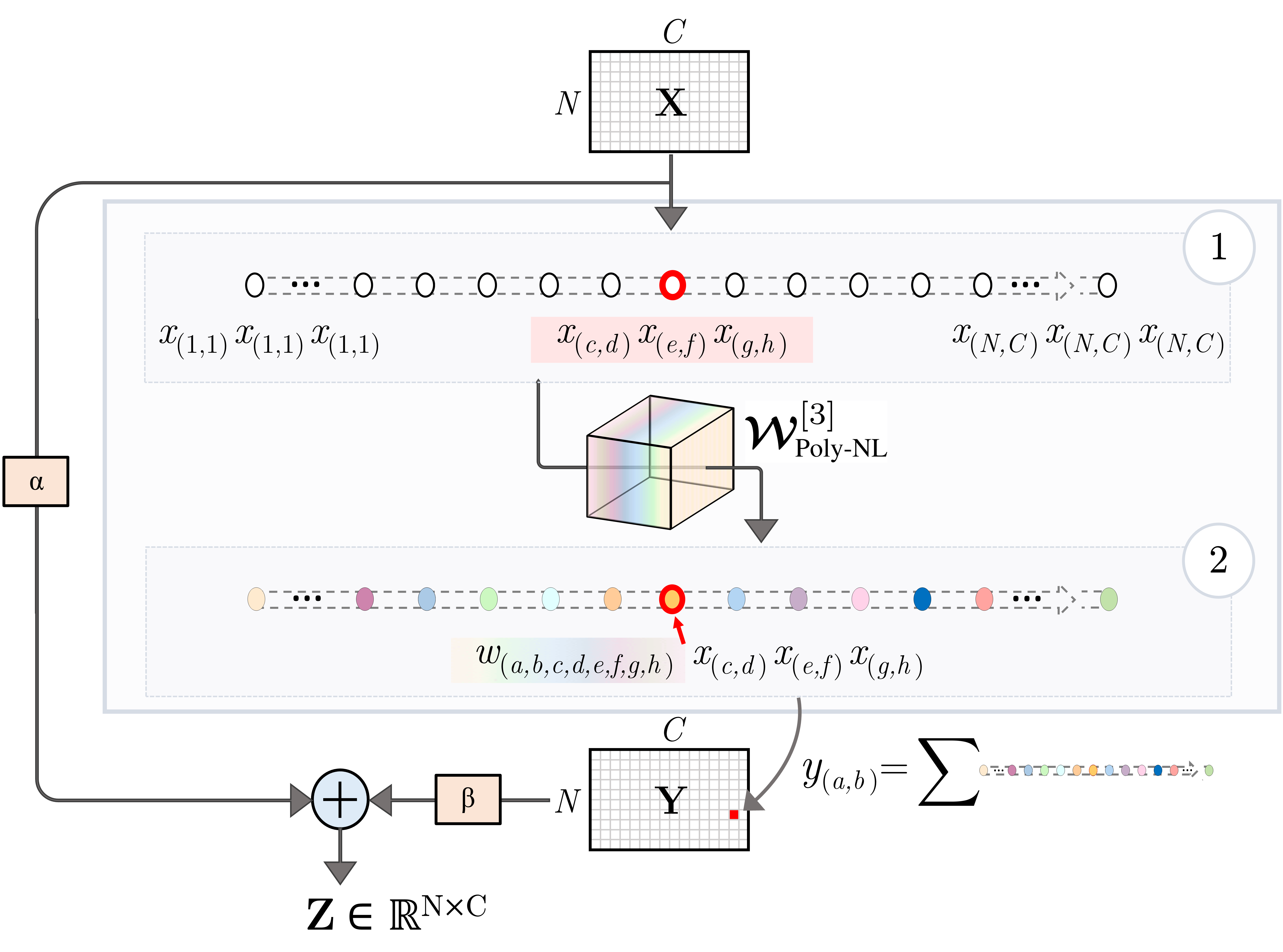}}   
    \vspace{-0.3cm}
    \caption{ \textbf{Two views of the Poly-NL block.} \textbf{ a)} Poly-NL as a non-local self-attention block for neural networks. The symbol $\odot$ denotes Hadamard products. Gray boxes represent convolutions of kernel size 1 and an averaging function over the rows. The output of the average pooling undergoes an expansion before the Hadamard multiplication.  \textbf{b)} Poly-NL as a $3^{rd}$ order polynomial module for neural networks. In the first box the space of $3^{rd}$ order interactions is represented as a line of $(NC)^3$ white dots, containing all possible triplets. The learnable parameters of $\mathbfcal{W}^{[3]}_{{\scalebox{.6}[.6]{Poly-NL}}} \in \mathbb{R}^{N \times C \times N \times C \times N \times C \times N \times C} $ weight each triplet $x_{(c,d)} x_{(e,f)} x_{(g,h)}$ by its importance $w_{(a,b,c,d,e,f,g,h)}$. This is depicted in the second box as a line of colored dots. The output element $y_{(a,b)}$ is the weighted summation of every triplet.  \vspace{-0.3cm}}
\label{fig:iiipoly}
\hspace*{-1cm}
\end{figure*}

\section{Related Work}
Multiplicative interactions~\cite{jayakumar2019multiplicative} can be found at the core of various machine learning models such as Bilinear layers, LSTM, and Higher‐order Boltzmann machines. In LSTM~\cite{hochreiter1997long, krause2016multiplicative}, element-wise products are used to fuse representations. In Bilinear layers~\cite{tenenbaum2000separating, carreira2012semantic, lin2015bilinear, yu2018hierarchical} feature maps of different networks get bilinearly combined together to capture pairwise interactions. In $k^{th}$‐order Boltzmann machines~\cite{memisevic2007unsupervised,memisevic2010learning,sejnowski1986higher} $k^{th}$ order multiplicative interactions are used to define the energy function. These \textit{high order interactions} capture the many possible ways in which the output can depend on the input. More recently, $\Pi-$nets~\cite{chrysos2020p} use polynomial expansions as a function approximator, replacing traditional activation functions with polynomials of the input vectors and use tensor decompositions~\cite{kolda2009tensor} to reduce the number of learnable parameters. 
 
 Multiplicative interactions are also crucial in the context of self-attention. Self-attention methods have been proposed as mechanisms to self-recalibrate feature maps and have been used either as replacement or addition to traditional residual blocks~\cite{he2016deep}.
 Complementary to our work, some of these methods accumulate contextual information into lightweight global-descriptors, either extrapolating a single scalar for each spatial position~\cite{wang2017residual}, channel~\cite{hu2018squeeze, cao2019gcnet}, channel and position~\cite{woo2018cbam} or region of space~\cite{li2019spatial}.
 
 Closer to our work, it is the idea of modeling non-local long-range dependencies among spatial positions. While this is not new in computer vision~\cite{buades2005non, lafferty2001conditional, dabov2007image} it is relatively recent in the context of neural networks architecture, in the form of ``Non-local" attention modules, capable to attend at the same time every element of the input~\cite{wang2018non, vaswani2017attention}. Examples of successful use of these modules can be found in natural language processing as well as in computer vision, where some form of self-attention has been used to achieve state-of-the-art performance in various problems as translation~\cite{ott2018scaling},  question answering~\cite{mohamed2019transformers}, classification~\cite{ramachandran2019stand, Bello_2019_ICCV}, segmentation~\cite{wang2020axial, carion2020end}, and video processing~\cite{wang2019edvr}, among others. While some works focused on extending the scope of Non-local blocks, by capturing channels' correlations~\cite{babiloni2020tesa, yue2018compact, fu2019dual} or considering multiple resolutions of the image~\cite{mei2020pyramid, dai2019second}, recent work has sparked a discussion on the scalability of these modules, and on how to overcome their  intrinsic efficiency limitations~\cite{tay2020efficient}.

Existing solutions focus on increasing the efficiency of the similarity operator, for example by reducing the number of positions attended~\cite{child2019generating, zhang2020dynamic} or using low dimensional latent spaces~\cite{choromanski2020rethinking, zhang2019latentgnn, chen2019graph, wang2020linformer} or on the computation order of the matrix-formula~\cite{shen2021efficient, katharopoulos2020transformers}. In this work, we propose an alternative solution to this problem.  We introduce a faster reformulation of the Non-Local block by framing non-local dependencies as $3^{rd}$ order interactions. Our method can extract non-local dependencies using no matrix multiplication computed along the spatial dimension.

\section{Non-locality and high-order interactions\vspace{-0.1cm}}
We start by introducing notation and background, then proceed in formalizing the concept of $3^{rd}$ order interactions. Our goal is to accelerate Non-local blocks in a principled manner without losing the rich, long-range interactions that have proven successful in practice. 

\subsection{Background}
\label{sec:background}
\paragraph{Notation.}{We follow the notation of Kolda \etal as in~\cite{kolda2009tensor}. Vectors are denoted as lower-case bold letters (e.g. $\mathbf{x}$) and matrices as upper-case bold letters (e.g $\mathbf{X}$). The element $(i,j)$ of a matrix $\mathbf{X}$ can be indicated as $x_{(i,j)}$. Tensors are identified with bold Euler script letters (e.g. $\mathbfcal{X}$). The order of a tensor is the number of dimensions, also known as way or mode. Hadamard products are indicated using the symbol ``$\odot$". Given two tensors, we define their double-dot product as the tensor contraction with respect to the last two indices of the first one and the first two indices of the second one, identified with the bullet ``$\bullet$" symbol. In the case of a tensor $\mathbfcal{W} \in \mathbb{R}^{I_{1} \times I_{2} \dots \times I_{N \text{-} 1} \times I_{N}}$ and a matrix $\mathbf{X} \in \mathbb{R}^{I_{N \text{-} 1} \times I_{N}} $ their double-dot product is a tensor of order {N \textbf{-} 2}, i.e. $\mathbfcal{Y}=\mathbfcal{W} \bullet \mathbf{X}$ of dimension $I_{1} \times I_{2} \dots \times I_{N \text{-} 2}$. Specifically, in element-wise form, such double-dot product reads  
$$y_{(i_{1},\dots,i_{N \text{-} 2})} = \sum\limits_{i_{n}=1}^{I_{N}} \sum\limits_{i_{n \text{-} 1}=1}^{I_{N \text{-} 1}} w_{(i_{1},\dots,i_{N \text{-} 2},i_{n \text{-} 1},i_{n})} x_{(i_{n \text{-} 1},i_{n})}.$$ \vspace{-0.8cm}\\}

\paragraph{Non-local Block.}{A generic self-attention block for neural networks highlights relevant interactions in a feature map using a function $g$, designed to manipulate the input, and a function $f$, in charge of extracting similarities from it. In~\cite{wang2018non}, the authors introduce the ``Non-local block", a self-attention block used to highlight non-local long-range dependencies in the input. It operates on a folded feature map $\mathbf{X} \in \mathbb{R}^{N \times C}$ of $N$ spatial positions and $C$ channels and outputs a matrix $\mathbf{Z}$ of the same dimensionality 
\begin{equation}
   \mathbf{Z} = \mathbf{Y} + \mathbf{X}= f(\mathbf{X})g(\mathbf{X}) + \mathbf{X}
   \label{eqn:sa}
\end{equation}
  where  $f\colon\mathbb{R}^{N \times C} \to \mathbb{R}^{N \times N}$ is a pairwise function that calculates similarity for each pair of spatial positions, and $g\colon\mathbb{R}^{N \times C} \to \mathbb{R}^{N \times C}$, which has the form of a unary function computing a new representation for the input. In the case where $g(\mathbf{X})$ is a linear embedding and $f(\mathbf{X})$ is an embedded dot-product, the contribution of the self-attention to the output can be written as
\begin{equation}
    \mathbf{Y}^{NL} =
    (\mathbf{X}\mathbf{W_{\theta}\mathbf{W_{\phi}^{\top}}\mathbf{X}^{\top}})(\mathbf{X}\mathbf{W}_{g}) = \mathbf{X}\mathbf{W}_{f}\mathbf{X}^{\top}\mathbf{X}\mathbf{W}_{g}
    \label{eqn:nl1_matrix}
\end{equation}
where $\mathbf{W_{\theta}}, \mathbf{W_{\phi}}, \mathbf{W}_{g} $ are matrices of learnable parameters of dimension $C \times C$.
To produce the output $\mathbf{Y}$, the Non-local block computes the dot-product between a similarity matrix $(\mathbf{X}\mathbf{W}_{f}\mathbf{X}^{\top}) \in \mathbb{R}^{N \times N}$ and an embedding of the input $(\mathbf{X}\mathbf{W}_{g}) \in \mathbb{R}^{N \times C}$. This matrix multiplication recalibrates the features of the $n^{th}$ position via aggregating information from all the others. The pairwise function provides the similarity weights for the contribution of each position and uses a matrix multiplication along the N dimension. 
Such matrix multiplication on the $N$ dimension is at the core of the non-local processing but introduces a quadratic term in computation that makes the complexity of this module equal to $O(N^2)$.}
\vspace{-0.1cm}
\paragraph{Polynomials for Neural Networks.}{ Recently in~\cite{chrysos2020p}, the authors adopted polynomials as layers of neural networks. We follow their formulation of a polynomial function $P\colon\mathbb{R}^{N \times C} \to \mathbb{R}^{N \times C}$ such that $\mathbf{Y}=P(\mathbf{X})$, in which each element of the output matrix is expressed as a polynomial of all the input elements $x_{(i,j)}$. The output of the layer is formed as
\begin{equation}
    \mathbf{Y} = P(\mathbf{X}) = \sum_{d=1}^{D} \mathbfcal{W}^{[d]}  \prod_{j=1}^{d}  \bullet  \mathbf{X} + \mathbf{W^{[0]}}
    \label{eqn:general_poly}
\end{equation}
where $D$ is the order of the polynomial, $\mathbfcal{W}^{[d]} $ is the tensor of learnable parameters associated with a specific order $d$, and $\mathbf{W^{[0]}}$ is a bias matrix of learnable parameters. The order of the tensor $\mathbfcal{W}^{[d]} $ increases exponentially for higher-order polynomial terms, i.e. $\mathbfcal{W}^{[d]} \in \mathbb{R}^{N \times C \times \prod_{j=1}^{d} (N\times C)}$.} 

\subsection{ 3$\mathbf{^{rd}}$ order interactions}
To present our method, we start by describing $3^{rd}$ order interactions terms for a feature map $\mathbfcal{X}\in \mathbb{R}^{H \times W \times C}$. We consider its folding $\mathbf{X}\in \mathbb{R}^{N \times C}$, where the spatial dimensions have been grouped together $N=H\times W$. To capture all potential $3^{rd}$ order dependencies between $\mathbf{X}$'s elements, we consider their linear combination weighted with a set of learnable parameters. In other words, we isolate the $3^{rd}$ order term ($D=3$) of Eq.~\eqref{eqn:general_poly} by assuming $\mathbfcal{W}^{[0]}=0, \mathbfcal{W}^{[1]}=0, \mathbfcal{W}^{[2]}= 0$. Under the aforementioned assumptions, Eq.~\eqref{eqn:general_poly} becomes 
\begin{equation}
    \mathbf{Y}  = ( ( ( \mathbfcal{W}^{[3]}  \bullet \mathbf{X} ) \bullet  \mathbf{X}) \bullet  \mathbf{X} )
    \label{eqn:general_poly_3}
\end{equation}
where $\mathbfcal{W}^{[3]}$ is a tensor of order $8$ and dimension $\mathbb{R}^{N \times C \times N \times C \times N \times C \times N \times C}$. We can find all possible $3^{rd}$ order interactions, i.e. the multiplication of all possible triplets of the input's elements summed together, clearly highlighted in its element-wise formula 
\begin{equation}
        y_{(a,b)}=\sum_{c,e,g}^{N}\sum_{d,f,h}^{C} w_{3_{(a,b,c,d,e,f,g,h)}} x_{(c,d)} x_{(e,f)} x_{(g,h)}
      \label{eqn:general_poly_3_element-wise}
\end{equation}
As depicted in Eq.~\eqref{eqn:general_poly_3_element-wise}, in a $3^{rd}$ order polynomial each element of the output matrix, $y_{(a,b)}$, benefits from the contributions of every possible triplet $ x_{(c,d)} x_{(e,f)} x_{(g,h)}$, each weighted by its unique importance $w_{3_{(a,b,c,d,e,f,g,h)}}$. The use of $\mathbfcal{W}^{[3]}$ in its most general form would allow taking into account every possible pattern in the input but, at the same time, it would increase the number of parameters exponentially. A well-known problem in higher-order models~\cite{ sejnowski1986higher, chrysos2020p} is the number of parameters considered, which tends to be the most expensive part of their implementation. The number of the parameters to determine in Eq.~\eqref{eqn:general_poly} depends on the order of the polynomial and, even without considering orders lower than D, the parameters required are $(NC)^{D}$ (for instance, the use of $D=3$ on an input $1024 \times 196$ will introduce nearly extra $10^{21}$ parameters).

Different approaches can be considered for reducing the number of parameters, for example by taking into account prior knowledge about the task or the nature of the input data~\cite{kolda2009tensor, memisevic2010learning}. One approach to reduce the complexity is by selecting only a limited subsets of all the possible combinations $ x_{(c,d)} x_{(e,f)} x_{(g,h)}$ exploiting a particular structure of the tensor $\mathbfcal{W}^{[3]}$. For example, assigning the same weight to a group of triplets will guarantee the same contribution for each of them, or imposing some zero-weights on one subsets of triplet will cancel their impact on the output $y_{(a,b)}$. These choices can be expressed in a formal way, which makes the format of the $\mathbfcal{W}^{[3]}$ tensor sparse, because some of the dimensions are constrained to be diagonal, or low-rank, since repeated values are used along some dimensions. 
The central idea of this paper is to factor the interaction tensor $\mathbfcal{W}^{[3]}$ in a particular way, to extract only a minimal subset of $3^{rd}$ order interactions from the input data.
In other words, the choice of such tensor allows its replacement with matrices of smaller size, implemented using only pre-existing building blocks for neural networks. 

\begin{figure*}[t]
\hspace*{-0.7cm}
  \subfloat[\centering]{\label{fig:big_c}\includegraphics[trim=0 0 10 0,clip,width=.356\linewidth, ]{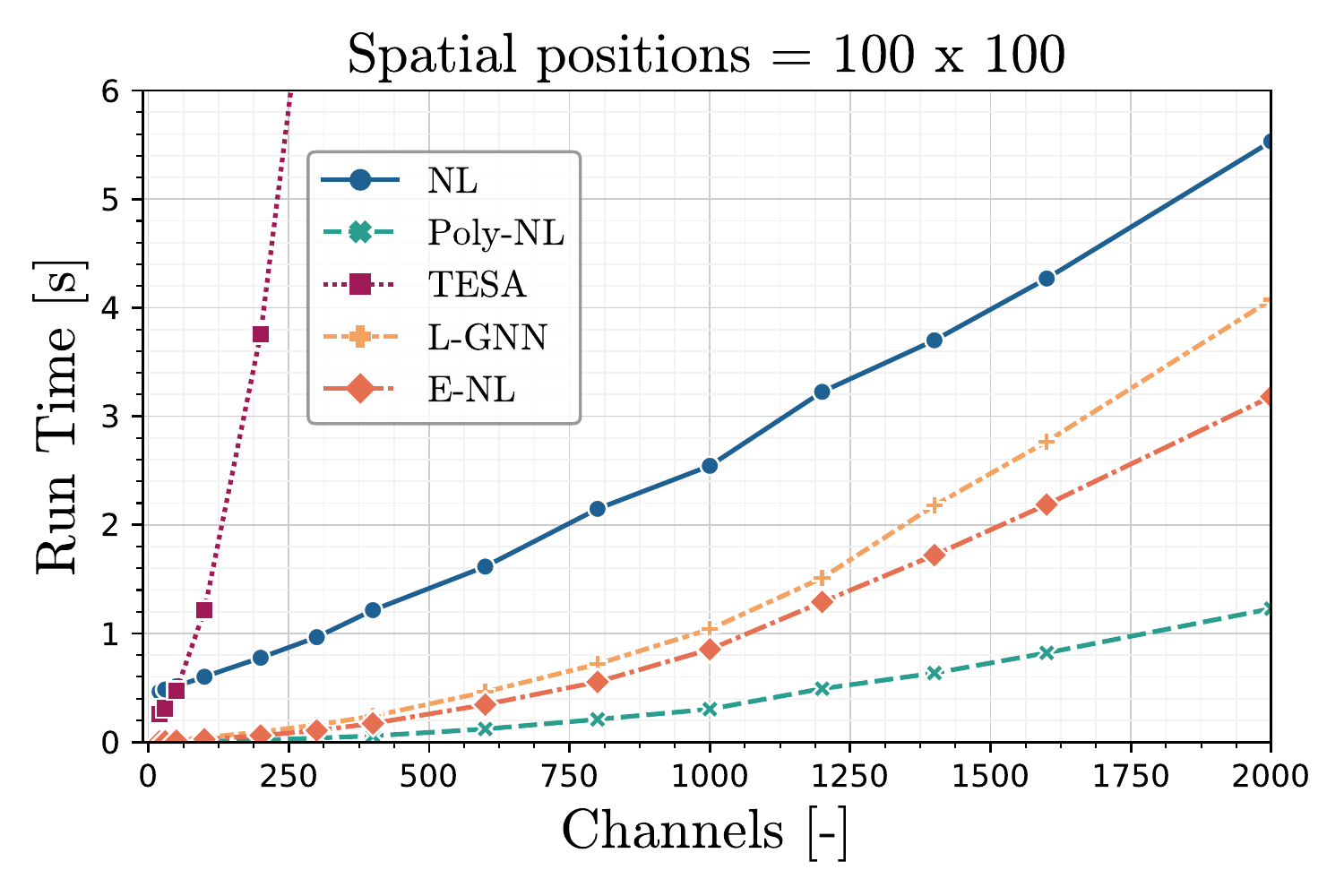}}   \subfloat[]{\label{fig:big_s}\includegraphics[trim=0 0 10 0,clip,width=.356\linewidth, ]{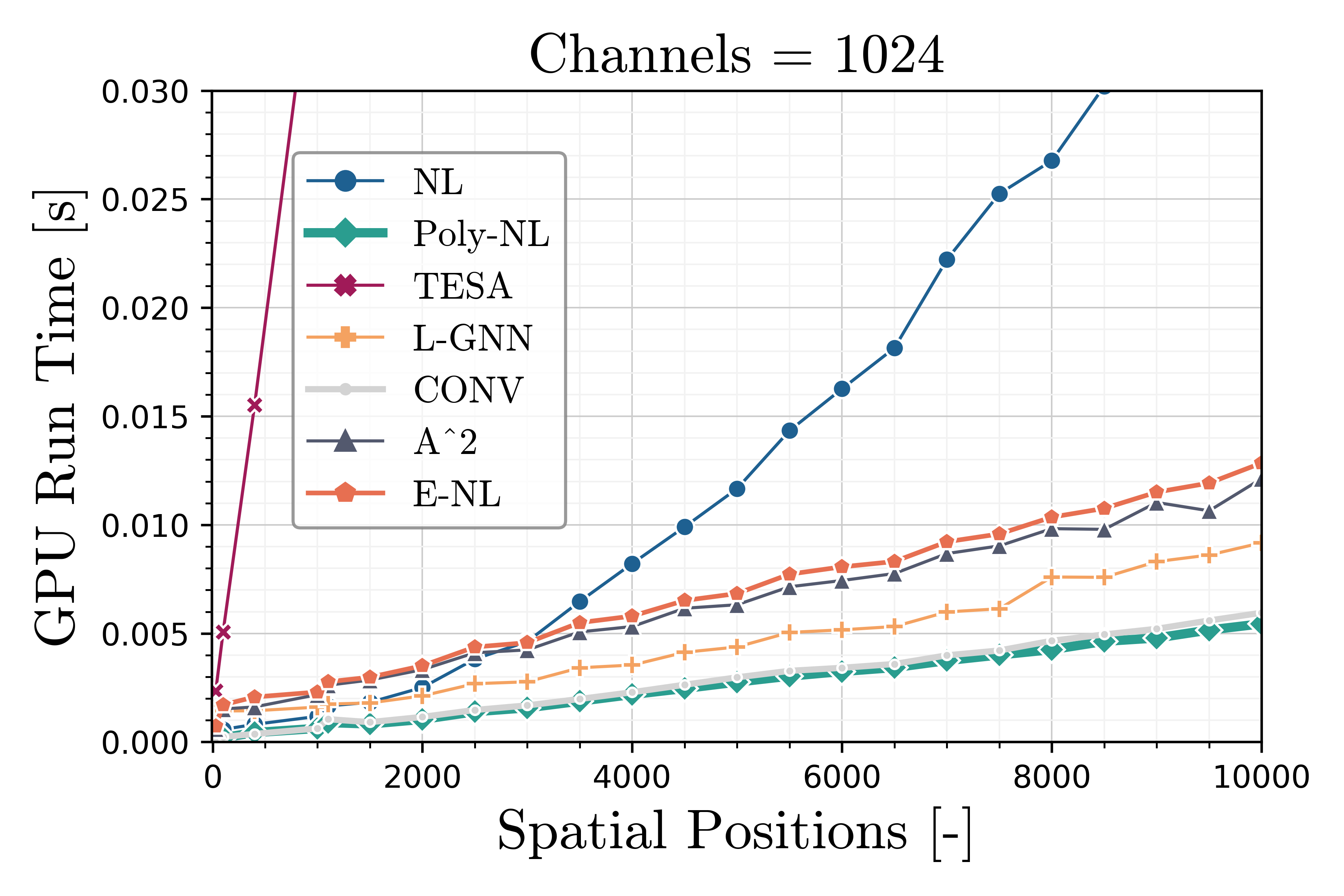}} 
  \subfloat[\centering  ]{\label{fig:small_s}\includegraphics[trim=0 0 10 0,clip,width=.356\linewidth, ]{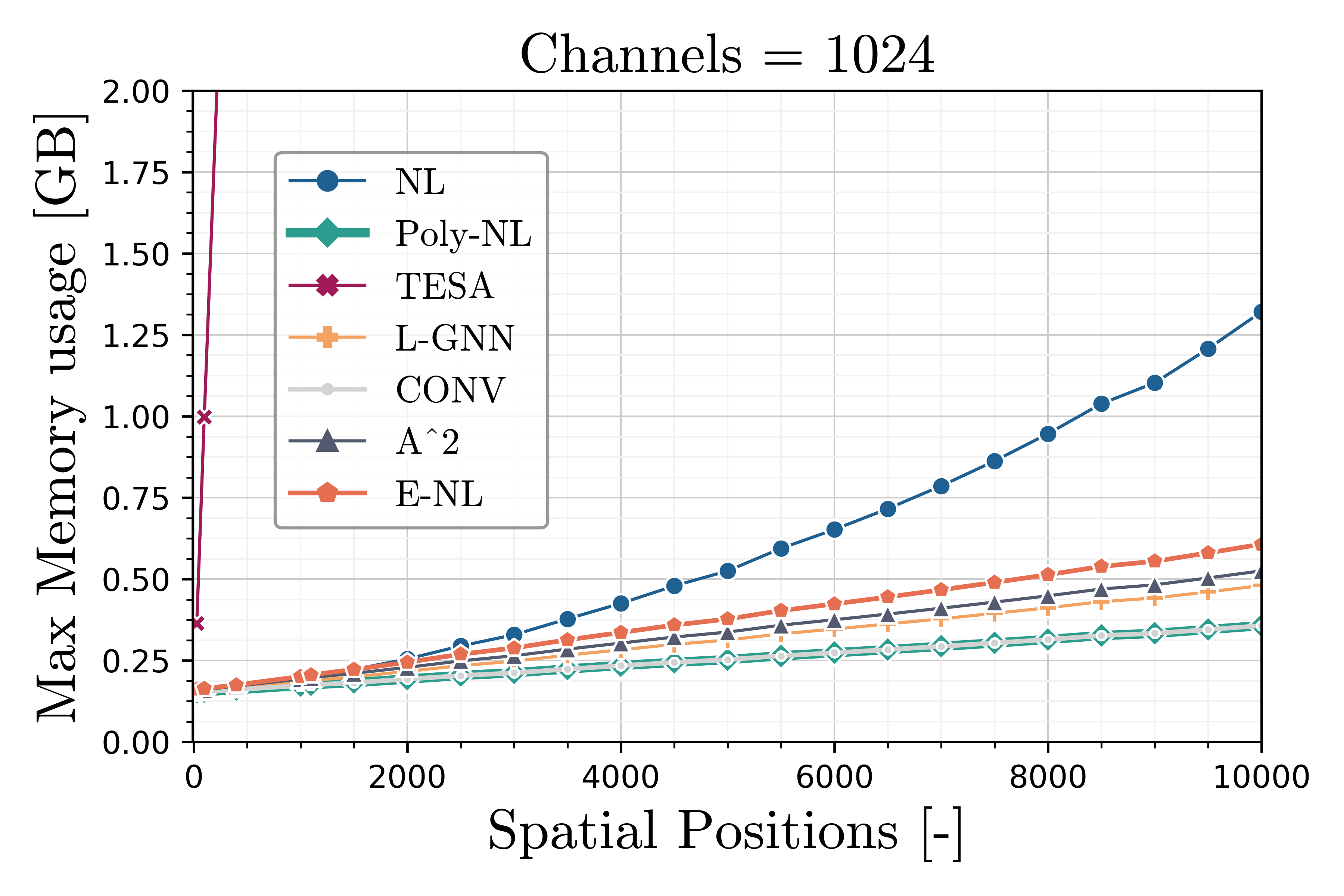}}   \vspace{-0.25cm}
  \caption{ \textbf{Runtime and Peak memory consumption} performance comparison between Poly-NL and other non-local methods executed on a CPU Intel(R) Core(TM) i9-9900X CPU (a) and a RTX2080 GPU (b,c). 
  Poly-NL exhibits lower computational overhead than competing methods, which is of importance with an increasing number of spatial positions. \vspace{-0.1cm}}
  \label{Fig:efficiency}
\hspace*{-1cm}
\end{figure*}

\begin{table*}[h!]
           \centering
           \captionsetup[subtable]{position = below}
          \captionsetup[table]{position=top}

{\renewcommand{\arraystretch}{1.1}
\begin{tabular}{p{2.3cm}p{.9cm}p{.9cm}p{.9cm}p{.9cm}p{1.1cm}p{1.3cm}}
\specialrule{.1em}{.05em}{.05em}
Method & $AP_{box}$ & $AP_{box50}$ & $AP_{box75}$ & $AP_{mask}$ & $AP_{mask50}$ & $AP_{mask75}$  \\ \hline
MaskR-CNN             & 37.9       & 59.2      & 41.0      & 34.6     & 56.0      & 36.9      \\
 + Non-local          & 38.8       & 60.6      & 42.0      & \underline{35.4}     & \underline{57.3}      & \underline{37.7}       \\
 + TESA               & \textbf{39.5}       & \textbf{60.9}      & \textbf{43.1}      & 35.4     & 57.2      & 37.5       \\
 + Latent-GNN         & 38.9       & 60.4      & \underline{42.4}      & 35.3     & 57.3      & 37.4       \\ 
 + Efficient-NL       & 38.9       & 60.3      & 42.2      & 35.4     & 57.2      & 37.7       \\
 \hline
 + Poly-NL               & \underline{39.2}       & \underline{60.8}      & 42.2      & \textbf{35.4}     & \textbf{57.4}      & \textbf{37.6}       \\ 
\specialrule{.1em}{.05em}{.05em}
\end{tabular}
}
\vspace{-0.1cm}
   \caption{\textbf{Results} of Poly-NL and other Non-Local methods on Instance Segmentation-COCO. \vspace{-0.25cm}}
   \label{tab:coco}    
\end{table*}


\section{Method}
In this section, we characterize the set of $3^{rd}$ order interactions associated with non-local dependencies and propose a method that accesses them without the expensive computation of a similarity matrix. The proposed method ``Poly-NL" is a novel non-local block, capable of selecting the same interactions as a Non-local layer at a fraction of its original computational cost in both space and time. 
\subsection{Poly-NL layer}
As described in Section~\ref{sec:background}, a major drawback of the Non-local block is its complexity, which depends on the number of spatial positions as $O(N^2)$.
To address this drawback we propose Poly-NL, a non-local spatial self-attention module that avoids any matrix multiplications along dimension N.  Poly-NL takes in input a matrix $\mathbf{X} \in \mathbb{R}^{N \times C}$ and outputs a matrix of the same dimensionality $\mathbf{Z}$, that can be computed as $\mathbf{Z} = \alpha \mathbf{X} + \beta \mathbf{Y}^{\scalebox{.6}[.6]{Poly-NL}}$, with $\alpha$ and $\beta$ are learnable scalars. The matrix $\mathbf{Y}^{\scalebox{.6}[.6]{Poly-NL}}$ is the core of the Poly-NL layer and can be written as follows
\begin{equation}
\mathbf{Y}^{\scalebox{.6}[.6]{Poly-NL}} = (\Phi(\mathbf{X}\mathbf{W}_{1} \odot \mathbf{X}\mathbf{W}_{2} ) \odot \mathbf{X}) \mathbf{W}_{3},
\label{eqn:fast_nl}
\end{equation}
where $\Phi\colon\mathbb{R}^{N \times C} \to \mathbb{R}^{N \times C}$ is an average pooling on spatial positions followed by an expand function, $\mathbf{W}_{1},\mathbf{W}_{2}, \mathbf{W}_{3} \in \mathbb{R}^{C \times C}$ are matrices of learnable parameters and $\odot$ indicates element-wise multiplication. A visual depiction of the module is presented in Figure~\ref{fig:iiipoly}, Poly-NL is a layer that scales linearly with the dimension N (i.e. has a complexity of $O(N)$).

Noteworthily, Poly-NL extracts the same set of dependencies as the Non-local block but learns a different set of weights to process them. In order to connect the two formulations, we describe the set of interactions captured by these two blocks. 
The set of spatial interactions associated with Poly-NL is clearly highlighted in its element-wise formula
\begin{equation}
 y_{(a,b)}^{\scalebox{.6}[.6]{Poly-NL}}= \sum_{d,f,h}^{C}\sum_{e}^{N} \frac{1}{N} w_{1_{(h,d)}} w_{2_{(f,d)}} w_{3_{(d,b)}} x_{(a,d)} x_{(e,f)} x_{(e,h)} 
\label{eqn:fast_nl_element-wise}
\end{equation}

In Poly-NL, each element $y_{(a,b)}^{\scalebox{.6}[.6]{Poly-NL}}$ of the output matrix is computed using the contribution of a set of triplets $x_{(a,d)} x_{(e,f)} x_{(e,h)}$, weighted using the learnable parameters $w_{1_{(h,d)}} w_{2_{(f,d)}} w_{3_{(d,b)}}$. 

Analogously, we highlight the set of interaction captured by the non-local module of Eq.~\eqref{eqn:nl1_matrix}, by writing its element-wise formulation
\begin{equation}
y_{(a,b)}^{\scalebox{.6}[.6]{NL}} = \sum_{d,f,h}^{C}\sum_{e}^{N} w_{f_{(d,f)}} w_{g_{(h,b)}} x_{(a,d)} x_{(e,f)} x_{(e,h)}.  
\label{eqn:nl1_element-wise}
\end{equation}
In the Non-local block, each element of the output matrix $y_{(a,b)}^{\scalebox{.6}[.6]{NL}}$ is computed using the contribution of a set of triplets $x_{(a,d)} x_{(e,f)} x_{(e,h)}$, weighted using the learnable parameters $w_{f_{(d,f)}} w_{g_{(h,b)}}$. As visible from the comparison between the two formulas, Poly-NL and Non-Local block modules are closely connected. They both access the same set of triplets and optimize through backpropagation a set of learnable weights. Nevertheless, the two modules differ considerably in terms of computational efficiency. Poly-NL does not need to explicitly compute any pairwise-function and can be therefore viewed as a linear complexity alternative to the Non-Local blocks.  

Interestingly, both of these blocks are also special cases of $3^{rd}$ order polynomials of Eq.~\eqref{eqn:general_poly_3}.
The outputs of these blocks $\mathbf{Y}^{{\scalebox{.6}[.6]{NL}}}$ and $\mathbf{Y}^{{\scalebox{.6}[.6]{Poly-NL}}}$, can be equivalently computed using Eq.~\eqref{eqn:general_poly_3}, in which $\mathbfcal{W}^{[3]}_{{\scalebox{.6}[.6]{NL}}}$ and $\mathbfcal{W}^{[3]}_{{\scalebox{.6}[.6]{Poly-NL}}}$ are block-sparse, low-rank, and can be decomposed through smaller matrices (i.e. $\mathbf{W}_g \mathbf{W}_f$ and $\mathbf{W}_{1} \mathbf{W}_{2} \mathbf{W}_{3}$, respectively). 

\begin{table*}[h!]
\centering
\begin{tabular}{p{2.8cm}p{.9cm}p{.9cm}p{.9cm}p{.9cm}p{1.1cm}p{1.1cm}}
\specialrule{.1em}{.05em}{.05em}
Method & $AP_{box}$ & $AP_{box50}$ & $AP_{box75}$ & $AP_{mask}$ & $AP_{mask50}$ & $AP_{mask75}$  \\ \hline
MaskR-CNN             & 37.9       & 59.2      & 41.0      & 34.6     & 56.0      & 36.9      \\\hline 
w/ Poly-NL             &        &       &       &      &       &        \\
 + on Res3               & 38.6       & 60.1      & 41.5      & 35.2     & 56.9      & 37.4       \\ 
 + on Res4               & \underline{39.2}       & \underline{60.8}      & \underline{42.2}      & \underline{35.4}     & \underline{57.4}      & \underline{37.6}       \\ 
 + on Res5               & 38.7       & 60.6      & 41.9      & 35.2     & 57.2      & 37.3       \\
 + on Res345             & \textbf{39.8}       & \textbf{61.7}      & \textbf{43.2}      & \textbf{36.0}     & \textbf{58.4}     & \textbf{38.3}       \\ 
\specialrule{.1em}{.05em}{.05em}
\end{tabular}
\vspace{-0.1cm}
\caption{\textbf{Ablation study} of Poly-NL placement in MaskR-CNN for Instance Segmentation. Adding Poly-NL on different ResNet blocks yields changes in performance. An application of Poly-NL on all ResNet blocks provides the best results when compared to a sole application on a single block.\vspace{-0.1cm}}
\label{tab:coco_ablation}
\end{table*} 
\begin{table*}[t]
\centering
\captionsetup[subtable]{position = below}
\captionsetup[table]{position=top}
\begin{subtable}{.38\linewidth}
\centering
{\renewcommand{\arraystretch}{1.1}
\begin{tabular}{lcc}
\specialrule{.1em}{.05em}{.05em}
Method & Top-1 & Top-5  \\ \hline
ResNet-50     & 75.62 & 92.68 \\
+ Non-local  & 76.09 & 93.00 \\ 
+ TESA       & \textbf{76.49} & \underline{93.05}  \\
+ Latent-GNN  & 75.28 & 92.33  \\
+ Efficient-NL  & 75.86 & 93.02 \\ 
\hline
+ Poly-NL       & \underline{76.30} & \textbf{93.06} \\ 
\specialrule{.1em}{.05em}{.05em}
\end{tabular}
}
\caption{\textbf{Imagenet}}
\label{tab:imagenet}
\end{subtable}%
\begin{subtable}{.7\linewidth}
\centering
{\renewcommand{\arraystretch}{1.1}
\begin{tabular}{p{2.8cm}p{1.3cm}p{1.3cm}p{1.3cm}}
\specialrule{.1em}{.05em}{.05em}
Method & $Easy$ & $Medium$ & $Hard$  \\ \hline
ResNet-50       & 95.49 & 94.85 & 89.87 \\
+ Non-local    & 95.88 & 95.14 & 91.94 \\
+ TESA         & \underline{96.22} & \underline{95.61} & \underline{92.58} \\
+ Latent-GNN   & 96.00 & 95.31 & 92.49 \\
+ Efficient-NL   & 96.06 & 95.42 & 92.55 \\ 
\hline
+ Poly-NL        & \textbf{96.37} & \textbf{95.71} & \textbf{92.76}  \\
\specialrule{.1em}{.05em}{.05em}
\end{tabular}
}
\caption{\textbf{Face Detection}}
\label{tab:face}
\end{subtable} 
\vspace{-0.1cm}
\caption{\textbf{Results} of Non-Local variants for image classification on ImageNet and face detection on WIDER FACE. \vspace{-0.1cm}}
\label{tab:3}
\end{table*}

\subsection{Relation with other Non-local blocks}
The idea of decomposing higher-order tensors in smaller matrices is not new~\cite{chi2012tensors, memisevic2010learning}, but can be used to cast a new light on a series of popular self-attention models. Besides Non-local block and Poly-NL, other popular non-local variants can be framed as special cases of $3^{rd}$ order polynomials~\cite{babiloni2020tesa, zhang2019latentgnn, shen2021efficient}. We compare Poly-NL to these methods and discuss the advantages of our formulation in terms of computational efficiency.

Figure~\ref{Fig:efficiency} depicts the complexity overhead of various spatial Non-local blocks for different sizes of the input matrix $\mathbf{X}$. In the visualization, we examine both the number of spatial positions (Figures~\ref{fig:big_s} and~\ref{fig:small_s}) and the number of channels (as in Figure~\ref{fig:big_c}). The charts examine the performance of five different methods (TESA~\cite{babiloni2020tesa}, NL~\cite{wang2018non}, L-GNN~\cite{zhang2019latentgnn}, E-NL~\cite{shen2021efficient} and Poly-NL) and a showcase how the proposed solution is able to process inputs size otherwise unmanageable by other formulations. We report computational time on CPU (Figure~\ref{fig:big_c}) and GPU (Figure~\ref{fig:big_s}) as measure of time complexity and peak memory usage on GPU as indicator of space complexity (Figure~\ref{fig:small_s}). To ease  comparisons among methods, we include as baseline a regular convolution layer (CONV), where no attention mechanism is used. All benchmarks were executed considering a single layer of each method on identical hardware, under comparable implementations, and hyper-parameters (please check additional material for full-nets run-times and implementation details). For each method, the values shown in the charts are the median of 20 runs.

The TESA block of~\cite{babiloni2020tesa} proposes to integrate spatial correlations together with channels' dependencies by computing six matrix multiplications on the three different matricizations of the input tensor $\mathbfcal{X}$. This procedure increments the patterns captured by the self-attention but it is burdened by a very high computational complexity of $O(N^2)$. The Latent-GNN block of~\cite{zhang2019latentgnn}, given the input matrix $\mathbf{X} \in \mathbb{R}^{N \times C}$, proposes to use a latent representation $N \times d$ to extract long-range dependencies with $O(N d^2)$ complexity. The block uses matrix multiplications to compute a low-rank matrix $d \times d$, which captures latent space interactions, and a matrix $d \times C$, which captures its relation with the input channels. This method has a computational complexity that is linear with respect to the number of spatial positions $N$ but depends on the choice of the hyperparameter $d$ and a sequence of matrix dot-product multiplications to compute the output. Lastly, the ``Efficient Non-local block"~\cite{shen2021efficient} proposes to compute the original formula of Eq.~\eqref{eqn:nl1_matrix} from right to left. This procedure avoids the computation of pairwise-spatial similarities and makes the complexity linear with respect to $N$, but it still requires computing a sequence of two matrix dot-products multiplications to extract the output. 

As displayed in Figures~\ref{fig:big_s} and~\ref{fig:small_s}, increasing the number of spatial positions greatly impacts efficiency. Run times of TESA and NL, which both depend quadratically on $N$, quickly become impractical, even in cases where the input dimension is small. Efficient methods (E-NL, L-GNN, Poly-NL) scale better with increasing spatial positions. Nonetheless, our method holds a competitive advantage across all figures, due to its lack of any matrix dot-product multiplication on the spatial dimension $N$. As shown in Figure~\ref{fig:big_c}, the number of channels impact linearly the runtime performance of most methods, with the notable exception of TESA. Even in this case, our proposed method is performing significantly better than competing methods especially when the number of channels becomes significant. 

As visible on the figures, Poly-NL consistently outperforms existing competitors, and has an efﬁciency on par with a regular convolution layer (CONV) since by design it avoids the explicit computation of any attention matrix.


\begin{figure*}[!htp]
\captionsetup[subfigure]{labelformat=empty}
\subfloat{\includegraphics[width=0.14\linewidth]{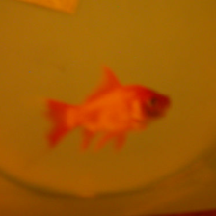}}
  \hspace*{0.08cm}
  \subfloat{\includegraphics[width=0.14\linewidth]{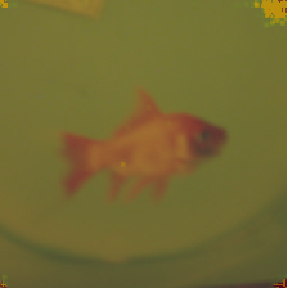}}
  \hspace*{0.08cm}
  \subfloat{\includegraphics[width=0.14\linewidth]{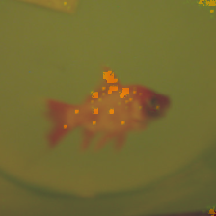}}
  \hspace*{0.08cm}
  \subfloat{\includegraphics[width=0.14\linewidth]{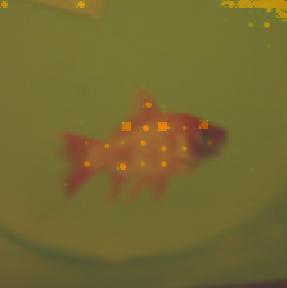}}
\hspace*{0.08cm}
  \subfloat{\includegraphics[width=0.14\linewidth]{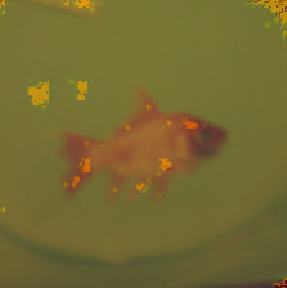}}
  \hspace*{0.08cm}
  \subfloat{\includegraphics[width=0.14\linewidth]{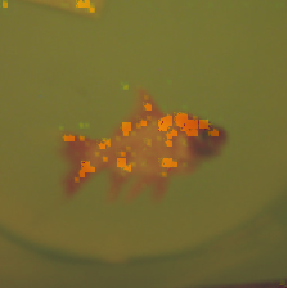}}  
  \hspace*{0.08cm}
  \subfloat{\includegraphics[width=0.14\linewidth]{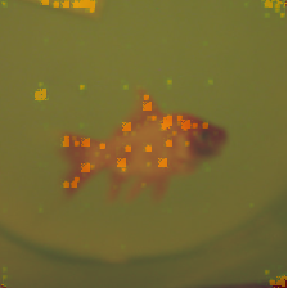}}
\\
\vspace{0.1cm}
\subfloat[Input Image]{\includegraphics[width=0.14\linewidth]{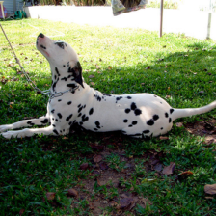}}
  \hspace*{0.08cm}
  \subfloat[ResNet-50]{\includegraphics[width=0.14\linewidth]{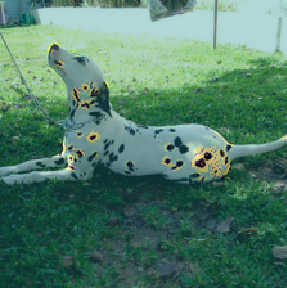}}
  \hspace*{0.08cm}
  \subfloat[+ Non Local]{\includegraphics[width=0.14\linewidth]{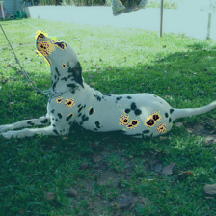}}
  \hspace*{0.08cm}
  \subfloat[+ Latent-GNN]{\includegraphics[width=0.14\linewidth]{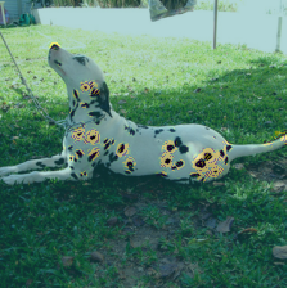}}
\hspace*{0.08cm}
  \subfloat[+ Efficient-NL]{\includegraphics[width=0.14\linewidth]{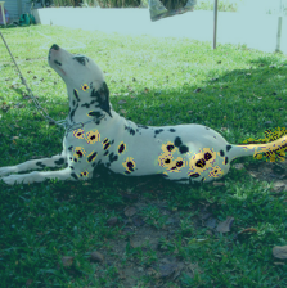}}
  \hspace*{0.08cm}
  \subfloat[+ TESA]{\includegraphics[width=0.14\linewidth]{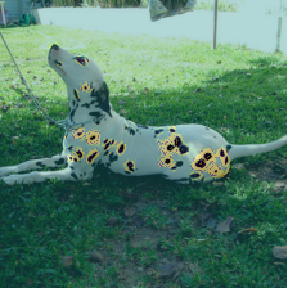}}  
  \hspace*{0.08cm}
  \subfloat[+ Poly-NL]{\includegraphics[width=0.14\linewidth]{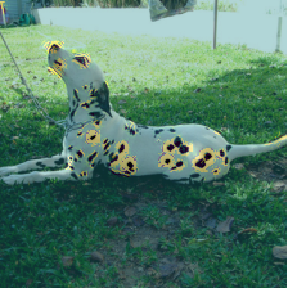}}
  \caption{ \textbf{Class saliency maps } of different methods. Grad-Cam~\cite{selvaraju2017grad} evaluates regions of the image which correspond to the class of interest. The use of non-local blocks helps to discriminate classes' characteristics.}
  \label{Fig:Grad_comparison}
\end{figure*}


\section{Experiments}
We evaluate the proposed method on three different tasks: object detection and instance segmentation on COCO~\cite{lin2014microsoft}, image classification on ImageNet~\cite{russakovsky2015imagenet}, and face detection on the WIDER FACE dataset~\cite{yang2016wider}. We provide empirical evidence that Poly-NL outperforms previously proposed Non-local neural networks while maintaining an optimal trade-off between efficiency and performance.

\subsection{Instance Segmentation on MS COCO}
We tested our method on object detection and instance segmentation, where the network processes an image and produces a pixel-pixel mask that identifies both the category and the instance for each object. We use the MS-COCO 2017 dataset~\cite{lin2014microsoft}, composed by 118k images as training set, 5k as validation set and 20k as test set, and the Mask R-CNN baseline of~\cite{he2017mask}. For all the experiments, we report the standard metrics of Average Precision $AP$, $AP_{50}$, and $AP_{75}$ for both bounding boxes and segmentation masks. The Mask R-CNN architecture is composed of a ResNet-FPN backbone for feature extraction followed by a stage that predicts class and box offsets. We trained with 8 Tesla V-100 GPUs and 2 images per GPU (effective batch size 16) using random horizontal flip as augmentation during training. We use an SGD solver with weight decay of 0.0001, momentum of 0.90, and an initial learning rate of 0.02. All models are trained for 26 epochs with learning rate steps are executed at epoch 16 and 22 with gamma 0.1. We used as backbones ResNet-50~\cite{he2016deep} architectures pre-trained on Imagenet.

Following prior work, we modify the Mask R-CNN backbone by adding one non-local layer right before the last residual block of Res4. This procedure highlights the capacity of the self-attention to boost features' representation and improve the quality of the candidate object bounding boxes. We compare our method against four different spatial self-attention layers, the original Non-local block of~\cite{wang2018non}, the efficient Latent-GNN variant of~\cite{zhang2019latentgnn}, the Efficient-NL of~\cite{shen2021efficient} and the recently proposed TESA~\cite{babiloni2020tesa}. For fair comparison, we report the results from our training, achieved using public available source codes and hyperparameters as provided by the respective authors. 

Quantitative results are summarized in Table~\ref{tab:coco}. When compared to the best performing method, TESA~\cite{babiloni2020tesa}, Poly-NL exhibits identical performance in $AP_{mask}$ and slightly lower accuracy for $AP_{box}$. However, we note that our proposed method is nearly $\times 10$ faster to compute than TESA at the given resolution. Moreover, compared to the Non-local layer~\cite{wang2018non} and its efficient variants Latent-GNN~\cite{zhang2019latentgnn} and Efficient-Net, our method improves performance by 0.3\% $\uparrow$  in $AP_{box}$ while keeping linear computational complexity.

We ablate the location we insert the proposed layer in MaskR-CNN and present our findings in Table~\ref{tab:coco_ablation}. We find that employing self-attention on any block of the ResNet backbone improves considerably the performance in both detection and segmentation. It appears that Res4 is the optimal block to insert Poly-NL into, since the numerical improvement across all metrics is consistent. At the same time, Table~\ref{tab:coco_ablation} shows that a combination of all ResNet blocks leads to the best performance (at most 1.2\% $\uparrow$  in $AP_{box}$  and 1.5\% $\uparrow$  in $AP_{mask}$). Although having a self-attention block at Res4 is preferable, the contribution of attention on multiple blocks outperforms the usage on a single module. These results suggest how complementary attention patterns can be captured at different network stages. 

\begin{figure*}[t]
\captionsetup[subfigure]{labelformat=empty}
\subfloat{\includegraphics[width=0.11\linewidth]{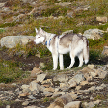}}
  \hspace*{0.3cm}
  \subfloat{\includegraphics[width=0.11\linewidth]{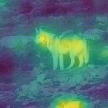}}
  \hspace*{0.3cm}
  \subfloat{\includegraphics[width=0.11\linewidth]{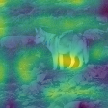}}
\hspace*{0.3cm}
  \subfloat{\includegraphics[width=0.11\linewidth]{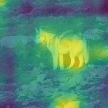}}
  \hspace*{0.4cm}
  \subfloat{\includegraphics[width=0.11\linewidth]{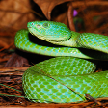}}
    \hspace*{0.3cm}
  \subfloat{\includegraphics[width=0.11\linewidth]{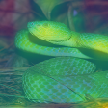}}
 \hspace*{0.3cm}
  \subfloat{\includegraphics[width=0.11\linewidth]{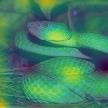}}
\hspace*{0.3cm}
  \subfloat{\includegraphics[width=0.11\linewidth]{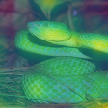}} \\
\vspace{0.1cm} 
\subfloat{\includegraphics[width=0.11\linewidth]{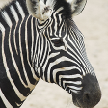}}
  \hspace*{0.3cm}
  \subfloat{\includegraphics[width=0.11\linewidth]{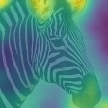}}
  \hspace*{0.3cm}
  \subfloat{\includegraphics[width=0.11\linewidth]{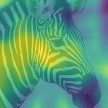}}
\hspace*{0.3cm}
  \subfloat{\includegraphics[width=0.11\linewidth]{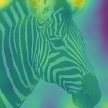}}
  \hspace*{0.4cm} 
  \subfloat{\includegraphics[width=0.11\linewidth]{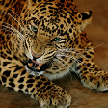}}
    \hspace*{0.3cm}
  \subfloat{\includegraphics[width=0.11\linewidth]{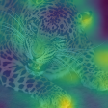}}
 \hspace*{0.3cm}
  \subfloat{\includegraphics[width=0.11\linewidth]{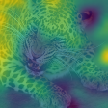}}
\hspace*{0.3cm}
  \subfloat{\includegraphics[width=0.11\linewidth]{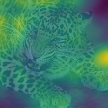}} \\
\vspace{0.1cm}
\subfloat{\includegraphics[width=0.11\linewidth]{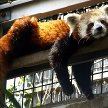}}
  \hspace*{0.3cm}
  \subfloat[\centering $\mathbf{X}$]{\includegraphics[width=0.11\linewidth]{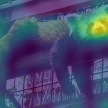}}
  \hspace*{0.3cm}
  \subfloat[\centering $\mathbf{Y}$]{\includegraphics[width=0.11\linewidth]{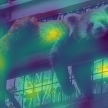}}
\hspace*{0.3cm}
  \subfloat[\centering $\mathbf{Z} = \mathbf{X} + \mathbf{Y}$  ]{\includegraphics[width=0.11\linewidth]{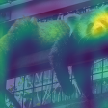}}
  \hspace*{0.4cm}
  \subfloat{\includegraphics[width=0.11\linewidth]{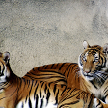}}
  \hspace*{0.3cm}
  \subfloat[\centering $\mathbf{X}$  ]{\includegraphics[width=0.11\linewidth]{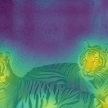}}
  \hspace*{0.3cm}
  \subfloat[\centering $\mathbf{Y}$]{\includegraphics[width=0.11\linewidth]{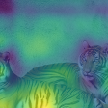}}
  \hspace*{0.3cm}
  \subfloat[\centering $\mathbf{Z} = \mathbf{X} + \mathbf{Y}$]{\includegraphics[width=0.11\linewidth]{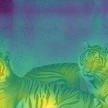}}
  \vspace{-0.1cm}
  \caption{ \textbf{Non-local dependencies} captured by Poly-NL. The norm of the extracted features per spatial location is visualized over the input image. The attention contribution $\mathbf{Y}$ learns patterns complementary to those captured by the input $\mathbf{X}$. The summation of the aforementioned quantities merges together the contribution of short and long-range spatial dependencies.\vspace{-0.1cm}}
  \label{Fig:Poly_NL_attention_visual}
\end{figure*}

\subsection{Face Detection on WIDER FACE}
We also apply our model to the task of face detection on the WIDER FACE dataset~\cite{yang2016wider}, which consists of $32,203$ images and $393,703$ face bounding boxes ($40\%$ training, $10\%$ validation, and $50\%$ testing) with a high degree of variability in scale, pose, expression, occlusion, and illumination. Compared to COCO~\cite{lin2014microsoft}, WIDER FACE~\cite{yang2016wider} contains more tiny and dense detection objects (\ie faces). 51\% of objects from COCO~\cite{lin2014microsoft} have the relative scale to the image below $0.11$, while for a similar proportion, 55\% of faces in WIDER FACE are less than $0.02$. In addition, 1\% of images in COCO have more than 30 objects, while there are 8\% images contains more than 30 faces in WIDER FACE and many images even include more than $150$ faces. Based on the detection rate of EdgeBox \cite{zitnick2014edge}, three levels of difficulty (\ie Easy, Medium, and Hard) are defined by incrementally incorporating hard samples. By using the evaluation metric of IoU $0.5$, we compare the Average Precision (AP) of the proposed method and other baselines on Easy, Medium and Hard subsets, respectively. 

Our experiments are implemented with PyTorch based on open source mmdetection \cite{chen2019mmdetection}. Inspired by RetinaNet \cite{lin2017focal}, we choose ResNet-50 \cite{he2016deep} as backbone and Feature Pyramid Network (FPN) \cite{lin2017feature} as neck to construct the feature extractor. The losses of classification and regression branches are focal loss \cite{lin2017focal} and DIoU loss \cite{zheng2020distance}, respectively. Following \cite{wang2018non}, we insert one Poly-NL block right before the last residual block of $c4$. To detect tiny faces, we tile three scales of anchors over each level of the FPN. The aspect ratio is set as $1.3$ and the IoU threshold for positive sampling matching is $0.35$. For augmentation during training, square patches are cropped and resized to $640 \times 640$ from the original image with a random scale. Then, photo-metric distortion and random horizontal flip with the probability of 0.5 are applied. We train the model by using SGD optimizer (momentum 0.9, weight decay 5e-4) with a batch size of $8\times 8$ on 8 Tesla V100 GPUs. The initial learning rate is set to $0.001$, linearly warms up for the first $3$ epochs, and decays by a factor of $10$ at $250$-th epoch and $350$-th epoch. All the models are trained with $400$ epochs from scratch without any pre-training. During testing, we only employ single scale inference with the short and long edge of image bounded by $1100$ and $1650$, respectively. As shown in Table~\ref{tab:face}, all attention modules can significantly improve the performance of face detection on WIDER FACE, indicating the effectiveness of context modeling (\ie capturing long-range dependencies among pixels). Besides, the proposed Poly-NL module consistently outperforms all other non-local layers across the three levels of difficulty, achieving the mAP of $92.76\%$ on the hard subset while being considerably faster than all competing methods.

\subsection{Classification on Imagenet}
We evaluated our method on the task of large-scale image classification, using Imagenet dataset \cite{russakovsky2015imagenet}, counting $1.28$M training images of $1000$ classes. For all the experiments, we modify a ResNet-50 architecture \cite{he2016deep} by inserting a self-attention module and then train from scratch with 8 GPUs for 90 epochs, using a batch size of 256 and an SGD optimizer with an initial learning rate of 0.1 and weight-decay as described in~\cite{goyal2017accurate}. Quantitative results are reported in table~\eqref{tab:imagenet} and show the Top-1 and the Top-5 accuracy for the evaluated methods. It is apparent that also in the classification task, where the goal is to provide a summary of the input, reasoning about spatial dependencies benefits greatly the accuracy. Poly-NL achieves the best performance on Top-5 accuracy, and on Top-1, outperforms significantly all other Non-Local neural networks with the exception of TESA~\cite{babiloni2020tesa}, which is very computationally demanding. 

Beyond quantitative results, Fig.~\ref{Fig:Grad_comparison} illustrates qualitative differences between different non-local variants. The visualization is produced with Gradient-weighted Class Activation
Mapping (Grad-CAM)~\cite{selvaraju2017grad}, a technique that highlights areas of high importance for image classification tasks. Our proposed method more accurately captures global context and salient features when compared to other Non-Local methods.

Fig.~\ref{Fig:Poly_NL_attention_visual} visualize the contribution of Poly-NL to the input's representation. 
The figure overlays the norm of different features on top of the input image. We report the input's feature map $\mathbf{X}$, the contribution of attention module $\mathbf{Y}$ and their summation $\mathbf{Z}$, i.e. output of the Poly-NL module. It is apparent that Poly-NL learns to contextualize the visual clues of the input with non-local dependencies. Our self-attention module can effectively recognize patterns that are complementary to those captured in the input and makes the features map aware of long-range dependencies. 

\section{Conclusions}
In this work, we cast the recently proposed Non-local block as a 3rd order polynomial in the form of multiplicative interactions between spatial locations on a grid. Based on this fact, we propose a novel and fast embodiment of Non-local layers named Poly-NL that can capture long-range dependencies with a complexity that scales linearly with the size of the input in both space and time. Poly-NL consistently outperforms other non-local networks on image recognition, instance segmentation, and face detection. 

{\small
\bibliographystyle{ieee_fullname}
\bibliography{Poly_NL}

\begin{thebibliography}{10}\itemsep=-1pt

\bibitem{babiloni2020tesa}
Francesca Babiloni, Ioannis Marras, Gregory Slabaugh, and Stefanos Zafeiriou.
\newblock Tesa: Tensor element self-attention via matricization.
\newblock In {\em Proceedings of the IEEE/CVF Conference on Computer Vision and
  Pattern Recognition}, pages 13945--13954, 2020.

\bibitem{Bello_2019_ICCV}
Irwan Bello, Barret Zoph, Ashish Vaswani, Jonathon Shlens, and Quoc~V. Le.
\newblock Attention augmented convolutional networks.
\newblock In {\em Proceedings of the IEEE/CVF International Conference on
  Computer Vision (ICCV)}, October 2019.

\bibitem{buades2005non}
Antoni Buades, Bartomeu Coll, and J-M Morel.
\newblock A non-local algorithm for image denoising.
\newblock In {\em 2005 IEEE Computer Society Conference on Computer Vision and
  Pattern Recognition (CVPR'05)}, volume~2, pages 60--65. IEEE, 2005.

\bibitem{cao2019gcnet}
Yue Cao, Jiarui Xu, Stephen Lin, Fangyun Wei, and Han Hu.
\newblock Gcnet: Non-local networks meet squeeze-excitation networks and
  beyond.
\newblock In {\em Proceedings of the IEEE/CVF International Conference on
  Computer Vision Workshops}, pages 0--0, 2019.

\bibitem{carion2020end}
Nicolas Carion, Francisco Massa, Gabriel Synnaeve, Nicolas Usunier, Alexander
  Kirillov, and Sergey Zagoruyko.
\newblock End-to-end object detection with transformers.
\newblock In {\em European Conference on Computer Vision}, pages 213--229.
  Springer, 2020.

\bibitem{carreira2012semantic}
Joao Carreira, Rui Caseiro, Jorge Batista, and Cristian Sminchisescu.
\newblock Semantic segmentation with second-order pooling.
\newblock In {\em European Conference on Computer Vision}, pages 430--443.
  Springer, 2012.

\bibitem{chen2019mmdetection}
Kai Chen, Jiaqi Wang, Jiangmiao Pang, Yuhang Cao, Yu Xiong, Xiaoxiao Li,
  Shuyang Sun, Wansen Feng, Ziwei Liu, Jiarui Xu, et~al.
\newblock Mmdetection: Open mmlab detection toolbox and benchmark.
\newblock {\em arXiv:1906.07155}, 2019.

\bibitem{chen2019graph}
Yunpeng Chen, Marcus Rohrbach, Zhicheng Yan, Yan Shuicheng, Jiashi Feng, and
  Yannis Kalantidis.
\newblock Graph-based global reasoning networks.
\newblock In {\em Proceedings of the IEEE/CVF Conference on Computer Vision and
  Pattern Recognition}, pages 433--442, 2019.

\bibitem{chi2012tensors}
Eric~C Chi and Tamara~G Kolda.
\newblock On tensors, sparsity, and nonnegative factorizations.
\newblock {\em SIAM Journal on Matrix Analysis and Applications},
  33(4):1272--1299, 2012.

\bibitem{child2019generating}
Rewon Child, Scott Gray, Alec Radford, and Ilya Sutskever.
\newblock Generating long sequences with sparse transformers.
\newblock {\em arXiv preprint arXiv:1904.10509}, 2019.

\bibitem{choromanski2020rethinking}
Krzysztof~Marcin Choromanski, Valerii Likhosherstov, David Dohan, Xingyou Song,
  Andreea Gane, Tamas Sarlos, Peter Hawkins, Jared~Quincy Davis, Afroz
  Mohiuddin, Lukasz Kaiser, David~Benjamin Belanger, Lucy~J Colwell, and Adrian
  Weller.
\newblock Rethinking attention with performers.
\newblock In {\em International Conference on Learning Representations}, 2021.

\bibitem{chrysos2020p}
Grigorios~G Chrysos, Stylianos Moschoglou, Giorgos Bouritsas, Yannis Panagakis,
  Jiankang Deng, and Stefanos Zafeiriou.
\newblock P-nets: Deep polynomial neural networks.
\newblock In {\em Proceedings of the IEEE/CVF Conference on Computer Vision and
  Pattern Recognition}, pages 7325--7335, 2020.

\bibitem{dabov2007image}
Kostadin Dabov, Alessandro Foi, Vladimir Katkovnik, and Karen Egiazarian.
\newblock Image denoising by sparse 3-d transform-domain collaborative
  filtering.
\newblock {\em IEEE Transactions on image processing}, 16(8):2080--2095, 2007.

\bibitem{dai2019second}
Tao Dai, Jianrui Cai, Yongbing Zhang, Shu-Tao Xia, and Lei Zhang.
\newblock Second-order attention network for single image super-resolution.
\newblock In {\em Proceedings of the IEEE/CVF Conference on Computer Vision and
  Pattern Recognition}, pages 11065--11074, 2019.

\bibitem{fu2019dual}
Jun Fu, Jing Liu, Haijie Tian, Yong Li, Yongjun Bao, Zhiwei Fang, and Hanqing
  Lu.
\newblock Dual attention network for scene segmentation.
\newblock In {\em Proceedings of the IEEE/CVF Conference on Computer Vision and
  Pattern Recognition}, pages 3146--3154, 2019.

\bibitem{goyal2017accurate}
Priya Goyal, Piotr Doll{\'a}r, Ross Girshick, Pieter Noordhuis, Lukasz
  Wesolowski, Aapo Kyrola, Andrew Tulloch, Yangqing Jia, and Kaiming He.
\newblock Accurate, large minibatch sgd: Training imagenet in 1 hour.
\newblock {\em arXiv preprint arXiv:1706.02677}, 2017.

\bibitem{he2017mask}
Kaiming He, Georgia Gkioxari, Piotr Doll{\'a}r, and Ross Girshick.
\newblock Mask r-cnn.
\newblock In {\em Proceedings of the IEEE international conference on computer
  vision}, pages 2961--2969, 2017.

\bibitem{he2016deep}
Kaiming He, Xiangyu Zhang, Shaoqing Ren, and Jian Sun.
\newblock Deep residual learning for image recognition.
\newblock In {\em Proceedings of the IEEE conference on computer vision and
  pattern recognition}, pages 770--778, 2016.

\bibitem{hochreiter1997long}
Sepp Hochreiter and J{\"u}rgen Schmidhuber.
\newblock Long short-term memory.
\newblock {\em Neural computation}, 9(8):1735--1780, 1997.

\bibitem{hu2018squeeze}
Jie Hu, Li Shen, and Gang Sun.
\newblock Squeeze-and-excitation networks.
\newblock In {\em Proceedings of the IEEE conference on computer vision and
  pattern recognition}, pages 7132--7141, 2018.

\bibitem{jayakumar2019multiplicative}
Siddhant~M Jayakumar, Wojciech~M Czarnecki, Jacob Menick, Jonathan Schwarz,
  Jack Rae, Simon Osindero, Yee~Whye Teh, Tim Harley, and Razvan Pascanu.
\newblock Multiplicative interactions and where to find them.
\newblock In {\em International Conference on Learning Representations}, 2019.

\bibitem{katharopoulos2020transformers}
Angelos Katharopoulos, Apoorv Vyas, Nikolaos Pappas, and Fran{\c{c}}ois
  Fleuret.
\newblock Transformers are rnns: Fast autoregressive transformers with linear
  attention.
\newblock In {\em International Conference on Machine Learning}, pages
  5156--5165. PMLR, 2020.

\bibitem{kolda2009tensor}
Tamara~G Kolda and Brett~W Bader.
\newblock Tensor decompositions and applications.
\newblock {\em SIAM review}, 51(3):455--500, 2009.

\bibitem{krause2016multiplicative}
Ben Krause, Liang Lu, Iain Murray, and Steve Renals.
\newblock Multiplicative lstm for sequence modelling.
\newblock {\em arXiv preprint arXiv:1609.07959}, 2016.

\bibitem{lafferty2001conditional}
John~D. Lafferty, Andrew McCallum, and Fernando C.~N. Pereira.
\newblock Conditional random fields: Probabilistic models for segmenting and
  labeling sequence data.
\newblock In {\em Proceedings of the Eighteenth International Conference on
  Machine Learning}, pages 282--289. Morgan Kaufmann Publishers Inc., 2001.

\bibitem{li2019spatial}
Xiang Li, Xiaolin Hu, and Jian Yang.
\newblock Spatial group-wise enhance: Improving semantic feature learning in
  convolutional networks.
\newblock {\em arXiv preprint arXiv:1905.09646}, 2019.

\bibitem{lin2017feature}
Tsung-Yi Lin, Piotr Doll{\'a}r, Ross Girshick, Kaiming He, Bharath Hariharan,
  and Serge Belongie.
\newblock Feature pyramid networks for object detection.
\newblock In {\em CVPR}, 2017.

\bibitem{lin2017focal}
Tsung-Yi Lin, Priya Goyal, Ross Girshick, Kaiming He, and Piotr Doll{\'a}r.
\newblock Focal loss for dense object detection.
\newblock In {\em ICCV}, 2017.

\bibitem{lin2014microsoft}
Tsung-Yi Lin, Michael Maire, Serge Belongie, James Hays, Pietro Perona, Deva
  Ramanan, Piotr Doll{\'a}r, and C~Lawrence Zitnick.
\newblock Microsoft coco: Common objects in context.
\newblock In {\em European conference on computer vision}, pages 740--755.
  Springer, 2014.

\bibitem{lin2015bilinear}
Tsung-Yu Lin, Aruni RoyChowdhury, and Subhransu Maji.
\newblock Bilinear cnn models for fine-grained visual recognition.
\newblock In {\em Proceedings of the IEEE international conference on computer
  vision}, pages 1449--1457, 2015.

\bibitem{liu2019spatially}
Chih-Ting Liu, Chih-Wei Wu, Yu-Chiang~Frank Wang, and Shao-Yi Chien.
\newblock Spatially and temporally efficient non-local attention network for
  video-based person re-identification.
\newblock In {\em British Machine Vision Conference}, 2019.

\bibitem{luo2017understanding}
Wenjie Luo, Yujia Li, Raquel Urtasun, and Richard Zemel.
\newblock Understanding the effective receptive field in deep convolutional
  neural networks.
\newblock In D. Lee, M. Sugiyama, U. Luxburg, I. Guyon, and R. Garnett,
  editors, {\em Advances in Neural Information Processing Systems}, volume~29.
  Curran Associates, Inc., 2016.

\bibitem{mei2020pyramid}
Yiqun Mei, Yuchen Fan, Yulun Zhang, Jiahui Yu, Yuqian Zhou, Ding Liu, Yun Fu,
  Thomas~S Huang, and Honghui Shi.
\newblock Pyramid attention networks for image restoration.
\newblock {\em arXiv preprint arXiv:2004.13824}, 2020.

\bibitem{memisevic2007unsupervised}
Roland Memisevic and Geoffrey Hinton.
\newblock Unsupervised learning of image transformations.
\newblock In {\em 2007 IEEE Conference on Computer Vision and Pattern
  Recognition}, pages 1--8. IEEE, 2007.

\bibitem{memisevic2010learning}
Roland Memisevic and Geoffrey~E Hinton.
\newblock Learning to represent spatial transformations with factored
  higher-order boltzmann machines.
\newblock {\em Neural computation}, 22(6):1473--1492, 2010.

\bibitem{mohamed2019transformers}
Abdelrahman Mohamed, Dmytro Okhonko, and Luke Zettlemoyer.
\newblock Transformers with convolutional context for asr.
\newblock {\em arXiv preprint arXiv:1904.11660}, 2019.

\bibitem{ott2018scaling}
Myle Ott, Sergey Edunov, David Grangier, and Michael Auli.
\newblock Scaling neural machine translation.
\newblock In {\em Proceedings of the Third Conference on Machine Translation
  (WMT)}, 2018.

\bibitem{ramachandran2019stand}
Prajit Ramachandran, Niki Parmar, Ashish Vaswani, Irwan Bello, Anselm Levskaya,
  and Jon Shlens.
\newblock Stand-alone self-attention in vision models.
\newblock In H. Wallach, H. Larochelle, A. Beygelzimer, F. d\textquotesingle
  Alch\'{e}-Buc, E. Fox, and R. Garnett, editors, {\em Advances in Neural
  Information Processing Systems}, volume~32. Curran Associates, Inc., 2019.

\bibitem{russakovsky2015imagenet}
Olga Russakovsky, Jia Deng, Hao Su, Jonathan Krause, Sanjeev Satheesh, Sean Ma,
  Zhiheng Huang, Andrej Karpathy, Aditya Khosla, Michael Bernstein, et~al.
\newblock Imagenet large scale visual recognition challenge.
\newblock {\em International journal of computer vision}, 115(3):211--252,
  2015.

\bibitem{sejnowski1986higher}
Terrence~J Sejnowski.
\newblock Higher-order boltzmann machines.
\newblock In {\em AIP Conference Proceedings}, volume 151, pages 398--403.
  American Institute of Physics, 1986.

\bibitem{selvaraju2017grad}
Ramprasaath~R Selvaraju, Michael Cogswell, Abhishek Das, Ramakrishna Vedantam,
  Devi Parikh, and Dhruv Batra.
\newblock Grad-cam: Visual explanations from deep networks via gradient-based
  localization.
\newblock In {\em Proceedings of the IEEE international conference on computer
  vision}, pages 618--626, 2017.

\bibitem{shen2021efficient}
Zhuoran Shen, Mingyuan Zhang, Haiyu Zhao, Shuai Yi, and Hongsheng Li.
\newblock Efficient attention: Attention with linear complexities.
\newblock In {\em Proceedings of the IEEE/CVF Winter Conference on Applications
  of Computer Vision}, pages 3531--3539, 2021.

\bibitem{tay2020efficient}
Yi Tay, Mostafa Dehghani, Dara Bahri, and Donald Metzler.
\newblock Efficient transformers: A survey.
\newblock {\em arXiv preprint arXiv:2009.06732}, 2020.

\bibitem{tenenbaum2000separating}
Joshua~B Tenenbaum and William~T Freeman.
\newblock Separating style and content with bilinear models.
\newblock {\em Neural computation}, 12(6):1247--1283, 2000.

\bibitem{vaswani2017attention}
Ashish Vaswani, Noam Shazeer, Niki Parmar, Jakob Uszkoreit, Llion Jones,
  Aidan~N Gomez, Lukasz Kaiser, and Illia Polosukhin.
\newblock Attention is all you need.
\newblock {\em arXiv preprint arXiv:1706.03762}, 2017.

\bibitem{wang2017residual}
Fei Wang, Mengqing Jiang, Chen Qian, Shuo Yang, Cheng Li, Honggang Zhang,
  Xiaogang Wang, and Xiaoou Tang.
\newblock Residual attention network for image classification.
\newblock In {\em Proceedings of the IEEE conference on computer vision and
  pattern recognition}, pages 3156--3164, 2017.

\bibitem{wang2020axial}
Huiyu Wang, Yukun Zhu, Bradley Green, Hartwig Adam, Alan Yuille, and
  Liang-Chieh Chen.
\newblock Axial-deeplab: Stand-alone axial-attention for panoptic segmentation.
\newblock In {\em European Conference on Computer Vision}, pages 108--126.
  Springer, 2020.

\bibitem{wang2020linformer}
Sinong Wang, Belinda Li, Madian Khabsa, Han Fang, and Hao Ma.
\newblock Linformer: Self-attention with linear complexity.
\newblock {\em arXiv preprint arXiv:2006.04768}, 2020.

\bibitem{wang2019edvr}
Xintao Wang, Kelvin~CK Chan, Ke Yu, Chao Dong, and Chen Change~Loy.
\newblock Edvr: Video restoration with enhanced deformable convolutional
  networks.
\newblock In {\em Proceedings of the IEEE/CVF Conference on Computer Vision and
  Pattern Recognition Workshops}, pages 0--0, 2019.

\bibitem{wang2018non}
Xiaolong Wang, Ross Girshick, Abhinav Gupta, and Kaiming He.
\newblock Non-local neural networks.
\newblock In {\em Proceedings of the IEEE Conference on Computer Vision and
  Pattern Recognition}, pages 7794--7803, 2018.

\bibitem{woo2018cbam}
Sanghyun Woo, Jongchan Park, Joon-Young Lee, and In~So Kweon.
\newblock Cbam: Convolutional block attention module.
\newblock In {\em Proceedings of the European conference on computer vision
  (ECCV)}, pages 3--19, 2018.

\bibitem{yang2016wider}
Shuo Yang, Ping Luo, Chen-Change Loy, and Xiaoou Tang.
\newblock Wider face: A face detection benchmark.
\newblock In {\em CVPR}, 2016.

\bibitem{yu2018hierarchical}
Chaojian Yu, Xinyi Zhao, Qi Zheng, Peng Zhang, and Xinge You.
\newblock Hierarchical bilinear pooling for fine-grained visual recognition.
\newblock In {\em Proceedings of the European conference on computer vision
  (ECCV)}, pages 574--589, 2018.

\bibitem{yue2018compact}
Kaiyu Yue, Ming Sun, Yuchen Yuan, Feng Zhou, Errui Ding, and Fuxin Xu.
\newblock Compact generalized non-local network.
\newblock In {\em NeurIPS}, pages 6511--6520, 2018.

\bibitem{zhang2020dynamic}
Li Zhang, Dan Xu, Anurag Arnab, and Philip~HS Torr.
\newblock Dynamic graph message passing networks.
\newblock In {\em Proceedings of the IEEE/CVF Conference on Computer Vision and
  Pattern Recognition}, pages 3726--3735, 2020.

\bibitem{zhang2019latentgnn}
Songyang Zhang, Xuming He, and Shipeng Yan.
\newblock Latentgnn: Learning efficient non-local relations for visual
  recognition.
\newblock In {\em International Conference on Machine Learning}, pages
  7374--7383, 2019.

\bibitem{zheng2020distance}
Zhaohui Zheng, Ping Wang, Wei Liu, Jinze Li, Rongguang Ye, and Dongwei Ren.
\newblock Distance-iou loss: Faster and better learning for bounding box
  regression.
\newblock In {\em AAAI}, 2020.

\bibitem{zitnick2014edge}
C~Lawrence Zitnick and Piotr Doll{\'a}r.
\newblock Edge boxes: Locating object proposals from edges.
\newblock In {\em ECCV}, 2014.

\end{thebibliography}
}

\end{document}